\documentclass{egpubl}
\usepackage{eg2016}

\ConferencePaper        % uncomment for (final) Conference Paper
 \electronicVersion % can be used both for the printed and electronic version

\ifpdf \usepackage[pdftex]{graphicx} \pdfcompresslevel=9
\else \usepackage[dvips]{graphicx} \fi

\PrintedOrElectronic

\usepackage{t1enc,dfadobe}

\usepackage{egweblnk}
\usepackage{cite}
\newcommand{\eps}{\epsilon}

\newcommand{\lam}{\lambda}
\newcommand{\gam}{\gamma}

\newcommand{\sig}{\sigma}
\newcommand{\sigmax}{\sig_{\max}}
\newcommand{\sigmin}{\sig_{\min}}

\newcommand{\aj}[1]{{\textcolor{black}{#1}}}
\newcommand{\new}[1]{{\textcolor{black}{#1}}}
\newcommand{\mew}[1]{{\textcolor{black}{#1}}}

\usepackage[normalem]{ulem}

\usepackage{epsfig}
\usepackage{graphicx}
\usepackage{amsmath}
\usepackage{amssymb}
\usepackage{bm}
\usepackage{color}
\usepackage{wrapfig}
\usepackage{placeins}
\usepackage[ruled,vlined]{algorithm2e}
\usepackage{tabularx}
\title[Learning 3D Deformation of Animals from 2D Images]%
      {Learning 3D Deformation of Animals from 2D Images}
\author[A. Kanazawa \& S. Kovalsky \& R. Basri \& D. Jacobs]{Angjoo
  Kanazawa$^1$, Shahar Kovalsky$^2$, Ronen Basri$^2$ and David Jacobs$^1$\\ 
$^1$University of Maryland College Park \quad $^2$Weizmann Institute of Science
       }

\begin{document}

% \teaser{
%  \includegraphics[width=\linewidth]{eg_new}
%  \centering
%   \caption{New EG Logo}
% \label{fig:teaser}
% }
\maketitle

\begin{abstract}
Understanding how an animal can deform and articulate is essential for
a realistic modification of its 3D model. In this
paper, we show that such information can be learned from user-clicked
2D images and a template 3D model of the target animal. We present a volumetric deformation framework that produces a set of
new 3D models by deforming a template 3D model according to a set of
user-clicked images. Our framework is based on a novel locally-bounded
deformation energy, where every local region has its own \emph{stiffness} value that bounds how much distortion is allowed at that location. We jointly learn the local stiffness bounds as we deform the
template 3D mesh to match each user-clicked image. We show that
this seemingly complex task can be solved as a sequence of convex optimization problems. We demonstrate the effectiveness of our
approach on cats and horses, which are highly deformable and articulated animals. Our framework produces new 3D
models of animals that are significantly more plausible than methods
without learned stiffness. 
% We introduce the idea of local-stiffness into the  a geometric deformation approach 

\begin{classification} % according to http://www.acm.org/class/1998/
\CCScat{Computer Graphics}{I.3.5}{Computational Geometry and Object
  Modeling}{Geometric algorithms, languages, and systems}
% ; \CCScat{Image Processing and Computer Vision}{I.4.5}{Reconstruction}{Transform Methods}
\end{classification}

\end{abstract}

%-------------------------------------------------------------------------
\section{Introduction}
\begin{figure*}[t]
\begin{center}
\includegraphics[width=\linewidth,height=50mm]{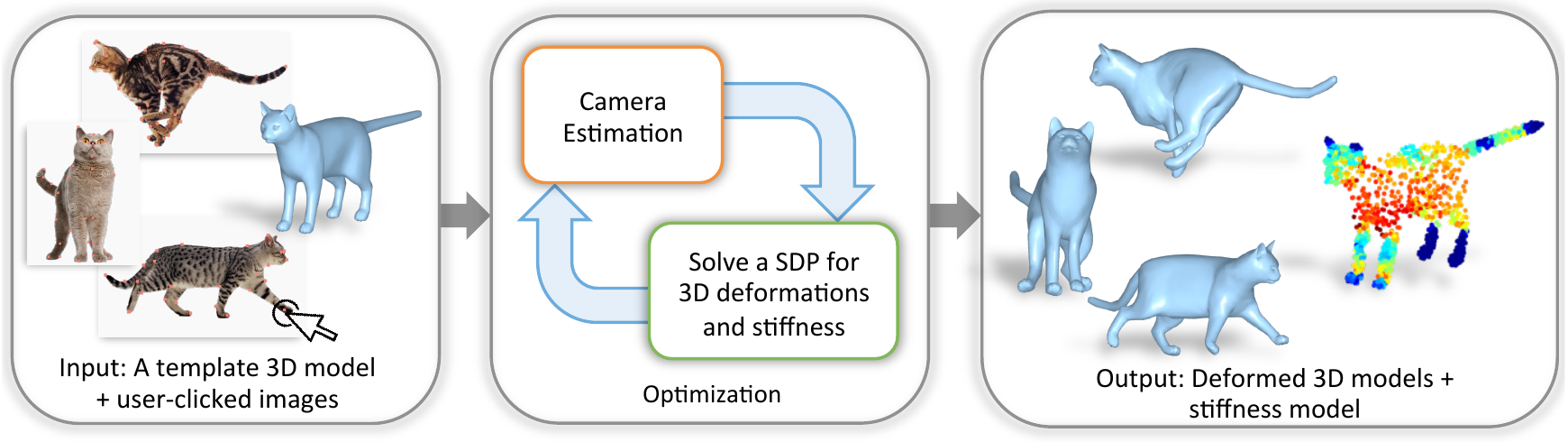}
   \caption{Overview. \new{Our inputs are a template 3D model of an
       animal and a set of images with user clicked 3D-to-2D point
       correspondences. The algorithm then alternates between solving
       for the camera viewpoint and the 3D deformations for all images. Our novel formulation allows us to  solve for the deformation for each image and the stiffness model
     of the animal jointly in a single semidefinite program (SDP). The
     outputs of our algorithm are a set of deformed 3D models and the
     stiffness model, which specifies the rigidity of every local region of the
     animal (red indicates high deformability and blue
     indicates rigidity).}}
\label{fig:pipeline}
\end{center}
\vspace{-1em}
\end{figure*}
Recent advances in computer vision and graphics have
enabled the collection of high-quality 3D models with tools such as
multi-view stereo \cite{Furukawa} and commercial depth-sensors
\cite{KinectFusion}. However, it is still difficult to obtain models of highly articulated and deformable objects like animals. % While online 3D databases are growing, models of animals are rare compared to rigid and man-made objects. 
Today, searching Turbosquid for  ``chair'' returns 24,929 results, while ``cat''
returns only 164 results. On the other hand, the Internet is filled with images of cats. The goal of our work is to create new 3D models of animals by modifying a template 3D model according to a set of user-clicked 2D images. The user clicks serve as positional constraints that guide the shape modification. 

In order to modify the shape realistically, we argue that it is
critical to understand how an animal can deform and articulate. For
example, looking at many images of cats shows that a cat's body may
curl up like a ball or twist and that its limbs articulate, but its
skull stays mostly rigid. Hence, when modifying a cat 3D model, we
should restrict the amount of deformation allowed around the skull,
but allow larger freedom around limb joints and the torso. % \aj{Our key
  % observation is that jointly considering many images of an anmial
  % can provide such local deformability information.} 

In this work, we propose a novel deformation framework that aims to
capture an animal-specific 3D deformation model from a set of annotated 2D images and a template 3D model. Our framework is inspired by the idea of \emph{local
  stiffness}, which specifies the amount of distortion allowed for a local region. Stiffness is used in 3D surface deformation methods to
model natural bending at joints and elastic deformations \cite{PopaSheffer,PriMo}. In previous methods, the stiffness is provided by users or learned from
a set of vertex-aligned 3D meshes in various poses
\cite{PopaSheffer}. \aj{Instead, we learn stiffness from user-clicked
  2D images by imposing sparsity; the idea is that large distortion is only allowed for those regions
  that require high deformation across many images.} To our knowledge,
our work is the first to learn stiffness of a 3D model from
annotated 2D images.

Figure \ref{fig:pipeline} shows an overview of our proposed
framework. Given a stock 3D cat mesh and target images of cats, a user provides 3D-to-2D point
correspondences by clicking key features in images. These are passed
on to the proposed algorithm, which simultaneously deforms the mesh to fit each cat's
pose and learns a cat-specific model of 3D deformation. In the end, we obtain new 3D models for each target image and a stiffness model that describes how cats may deform and articulate.

% \aj{We use user clicks for obtaining 2D-to-3D correspondences similar to recent works such as \cite{Vincente,Karsch,PrasadNemo}, since computing
%   automatic correspondences between different animals is still beyond
%   the state-of-the-art. An interesting follow-up could be to use our
%   models as a shape prior for the automatic detection of corresponding points.}

% \vspace{0.5cm}
\noindent\textbf{Contributions:} 
Our primary contribution is a deformation framework that learns an animal-specific model of local stiffness as it deforms the template model to match the user-clicked
2D-to-3D correspondences. Specifically,
\begin{enumerate}
\item We propose a locally bounded volumetric deformation energy that
  controls the maximal amount of distortion applied to local
  regions of the model using the recent optimization techniques of \cite{ContSingVal}. The bounds act as a local stiffness model of the
  animal, which we learn by imposing a L1 sparsity penalty. The final deformation is orientation preserving and has worst-case distortion guarantees. 
\item We show that both the deformation and the stiffness bounds can be solved jointly as a sequence of convex optimization problems.
\item We demonstrate the effectiveness of our framework on cats and horses, which are
challenging animals as they exhibit large degrees of  deformation and articulation.
\end{enumerate}
% Our experiments show that learning stiffness from multiple images allows for a more plausible deformation
% of the template 3D mesh. 
%-------------------------------------------------------------------------
\section{Related Work}
Works that modify a template 3D model to fit images can be roughly
divided into two categories: those that are class-agnostic, and those
that have a class-specific deformation prior learned from data or provided by users. Methods that do not use any class-specific prior
make use of strong image cues such as silhouettes
\cite{Vincente,GuanghuaStyle} or contour drawings
\cite{Kraevoy}. These approaches focus on fitting a single 3D model
into a single image, while we focus on learning a class-specific
prior as we modify the template 3D model to fit multiple images. Recently, Kholgade et al. introduced an exciting new photo editing tool that allows users to perform 3D
manipulation by aligning 3D stock models to 2D images
\cite{Kholgade}. Our approach \new{complements} this application, which is only demonstrated for rigid objects.

More closely related to our approach are works that make use of prior
knowledge on how the 3D model can change its shape. Many works assume
a prior is provided by users or artists in the form of kinematic skeletons
\cite{Guan,Akhter,hand,Ballan,MaoYe,Taylor14} or painted stiffness
\cite{PopaSheffer}. Since obtaining such priors from users is
expensive, many methods learn deformation models automatically from data \cite{BlanzVetter,Anguelov,SCAPE,ChenCipolla,deAguiar,Skeleton,Schaefer}. Anguelov et al. \cite{Anguelov} use a set of registered 3D range scans of human bodies in a variety of configurations to construct skeletons using graphical models. Blanz and Vetter \cite{BlanzVetter} learn a morphable model of human faces from 3D scans, where a 3D face is described by a linear combination of basis faces. Given a rough initial alignment, they fit the learned morphable models to images by restricting the model to the space spanned by the learned basis. Similarly \cite{SCAPE,Hasler} learn a statistical model of human bodies from a set of 3D scans. Popa et al. \cite{PopaSheffer} learn the material stiffness of animal meshes by analyzing a set of vertex-aligned 3D meshes in various poses. 

One of the biggest drawbacks in learning from 3D data is that it
requires a large set of registered 3D models or scans, which is
considerably more challenging to obtain compared to a set of
user-clicked photographs. All of these methods rely on 3D data with
the exception of Cashman et al. \cite{Cashman}. They learn a morphable
model of non-rigid objects such as dolphins  from annotated
2D images and a template 3D model. Our work is \new{complementary} to their
approach in that they focus on intra-class shape variation such as fat
vs thin dolphins, while we focus on deformations and articulations due
to pose changes. The use of a morphable model also makes their approach
not suitable for objects undergoing large articulations. 

Using 2D images requires camera parameters for projecting the deformed 3D models to image coordinates. Cashman et
al. \cite{Cashman} assume a rough camera initialization is provided by a user, but we estimate the camera parameters directly from
user-clicked 2D-to-3D correspondences. There are many works regarding
the estimation of camera parameters from image correspondences, and
their discussion is outside the scope of this paper. We refer the reader to \cite{HZ} for more details. 

% The problem of matching 3D models to 2D images dates all the way back to 60's with L.G. Roberts' attempt to fit a collection of polyhedral blocks to images \cite{Roberts}. More recently,
% Recently, Kholgade et al. presented an exciting new photo editing tool that allows users to perform 3D
% manipulation by aligning 3D stock models to 2D images \cite{Kholgade}. Our work compliments
% this application which is only demonstrated for rigid
% objects. 

There is a rich variety of mesh deformation techniques in the literature \cite{BotschSorkine,Zhou,PriMo,Sumner07,ARAP}. The main idea is to minimize some form of
deformation objective that governs the way the mesh is modified
according to user-supplied positional constraints. Common objectives
are minimization of the elastic energy \cite{shell} or preservation of local
differential properties \cite{lipman}. The solution can be constrained
to lie in the space of natural deformations, which are learned from
exemplar meshes \cite{SumnerPopovic,Sumner, Der,
  martin2011example,PopaSheffer}. Our approach is related to these
methods, except that we learn the space of deformations from a set of annotated 2D images. \cite{BotschSorkine} offers an excellent survey on linear surface
deformation methods. While simple and efficient to use, surface
deformation methods suffer from unnatural volumetric changes for
large deformations \cite{Zhou,PriMo}. Our work is based on a
volumetric representation, which we discuss in detail in the next
section. 

% The problem of skeleton extraction from a single mesh is an ill-posed problem as there is little information about the
% articulation. \cite{skinningcourse} presents a thorough discussion
% of these challenges and recent developments in graphics for automatic
% skinning. Furthermore, it is not always clear what is the right
% skeleton model when the object deforms in an elastic manner. Our model is more flexible than skeleton models and can capture both articulation and deformation.

%-------------------------------------------------------------------------

%-------------------------------------------------------------------------
\section{Problem statement and background}
\label{sec:bg}
We consider the problem of modifying a template 3D mesh of an animal
according to a set of user-clicked photographs of the target animal. \new{Our
goal is to produce plausible 3D models guided by the annotated images, not necessarily obtaining precise 3D reconstructions of the
images.} In particular, given
a sparse set of 2D-to-3D correspondences obtained from user-clicks, we
wish to solve for a set of class-specific 3D deformations that
faithfully fit the image annotations. 

More formally, we are given a 3D template model, represented by a surface mesh $\mathbf{S} \subset \mathbb{R}^3$ as well as $N$ images of class instances $I^1,\ldots,I^N$. Each image is associated with  a sparse set of user prescribed point correspondences to the 3D model; namely, the $i$'th image $I^i$ comes with pairs $\{(\bm{x}^i_k,\bm{p}^i_k)\}$ relating the surface point $\bm{x}^i_k\in\mathbf{S}$ to a 2D image location $\bm{p}^i_k\in\mathbb{R}^2$. Our goal is to leverage the $N$ annotated images to learn a
deformation model $\mathcal{D}$ capturing the possible deformations
and articulations of the object class. In particular, for each image $I^i$ we wish to find a
deformation $\Phi^i\in\mathcal{D}$ that maps its 3D landmark points
$\{\bm{x}^i_k\}$  to their corresponding image points $\{\bm{p}^i_k\}$ once projected to the image plane; namely, satisfying
\begin{equation}
\label{eqn:pos_const_1}
\begin{bmatrix}
\bm{p}^i_k\\
1
\end{bmatrix}
=
\Pi^i
\begin{bmatrix}
\Phi^i(\bm{x}^i_k)\\
1
\end{bmatrix},
\end{equation}
where $\Pi^i\in \mathbb{R}^{3 \times 4}$ is the camera projection
matrix for the $i$'th image. In what follows we assume weak
perspective projection, which \new{is an orthographic projection followed by scaling of the $x$ and $y$ coordinates:
  \begin{equation}
    \label{eqn:camera}
\Pi =
\begin{bmatrix}
  \alpha_x & & \\
  & \alpha_y & \\
& & 1
\end{bmatrix}
\begin{bmatrix}
  \mathbf{r}_1 & t_1\\
  \mathbf{r}_2 & t_2\\
\mathbf{0} & 1
\end{bmatrix}.
% \begin{bmatrix}
% 1 & 0 & 0 & 0\\
% 0 & 1 & 0 & 0\\
% 0 & 0 & 0 & 1\\
% \end{bmatrix}
% \begin{bmatrix}
% R & t
% \end{bmatrix}.
  \end{equation}
$\mathbf{r}_1$ and $\mathbf{r}_2$ are the first two rows of the object
rotation matrix, $t_1, t_2$ are the first two coordinates of the
object translation, and $\frac{\alpha_x}{\alpha_y}$ specifies the
camera aspect ratio. Its parameters can be solved in a least squares
approach given six or more 3D-to-2D point correspondences. Please see
\cite{HZ} for more information. Note that perspective projection may
be similarly handled.}

\subsection{Parameterized deformation model}
\label{sec:param}
\begin{figure}[t]
\begin{center}
  \includegraphics[width=\linewidth,height=35mm]{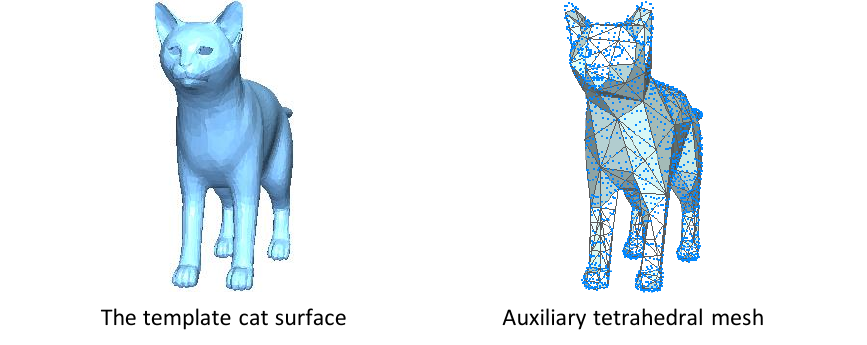}
\end{center}
\caption{A template 3D surface and its auxiliary tetrahedral
mesh with surface vertices shown in blue dots.}
\label{fig:surftet}
\vspace{-1.5em}
\end{figure}

We parameterize the deformations of the surface model $\mathbf{S}$ by introducing an auxiliary tetrahedral mesh enclosed within the surface, $\mathbf{M} = (\mathbf{V},
\mathbf{T})$, where $\mathbf{V} \in \mathbb{R}^{3
  \times n}$ is a matrix of $n$ coarse vertex coordinates and
$\mathbf{T} = \{t_j\}_{j=1}^m$ is a set of $m$ tetrahedra (tets). Every surface point $\bm{x}\in\mathbf{S}$ can then be written as a
linear combination of the vertices $\mathbf{V}$. In particular, for
the landmark points we set $\bm{x}^i_k=\mathbf{V}\bm\alpha^i_k$, where
$\bm\alpha^i_k \in \mathbb{R}^{n}$ is a coefficient vector computed by
linear moving least squares \cite{MLS}. Figure \ref{fig:surftet} shows the surface and the
tetrahedral mesh of a template cat model. The use of a tetrahedral
mesh introduces a notion of volume to the model making it more robust at preserving volumetric detail \cite{Edilson,Zhou}. 

Deformations of $\mathbf{M}$ thereby induce deformations of the surface $\mathbf{S}$. Specifically, we shall consider continuous piece-wise linear (CPL) maps $\Phi:\mathbf{M} \rightarrow \mathbb{R}^3$, whereby the deformation, restricted to the $j$'th tet, is defined by the affine map $\bm{v} \mapsto A_j\bm{v} + \bm{t}_j$. $\Phi$ maps the vertices $\mathbf{V}$ to new locations $\mathbf{U} \in \mathbb{R}^{3 \times n}$. In fact, $\Phi$ is uniquely determined by the new vertex locations $\mathbf{U}$; for the $j$'th tet, the following full rank linear system holds
\begin{equation}
\left(  \begin{smallmatrix}
    \bm u_{j1}& \bm u_{j2}& \bm u_{j3} &\bm u_{j4}  
  \end{smallmatrix} \right)
  =
  \begin{bmatrix}
    A_j & \bm{t}_j   
  \end{bmatrix}
  \left(\begin{smallmatrix}
    \bm v_{j1} & \bm v_{j2} &\bm v_{j3} & \bm v_{j4} \\
    1& 1& 1& 1
  \end{smallmatrix}\right), \label{eqcont0}
\end{equation}
where $\bm v_{j\cdot}$ and $\bm u_{j\cdot}$ are its four vertices in the original and the deformed mesh respectively. 
% \begin{figure}[to]
% \begin{center}
%   \begin{subfigure}{0.48\linewidth}
%     \begin{center}
%       \resizebox{0.5\columnwidth}{!}{\includegraphics{figures//cat_surf.png}}
%     \caption{\small A template cat surface}
%     \label{fig:surf}
%     \end{center}
%   \end{subfigure}  
%   \begin{subfigure}{0.48\linewidth}
%     \begin{center}
%       \resizebox{0.5\columnwidth}{!}{\includegraphics{figures//cat_tet.png}}
%       \caption{\small Tetrahedral mesh}
%       \label{fig:tet}
%     \end{center}
%   \end{subfigure}
% \end{center}
% \vspace{-1em}
% \end{figure}

\begin{figure}[b]
\centerline{\resizebox{\columnwidth}{!}{\includegraphics{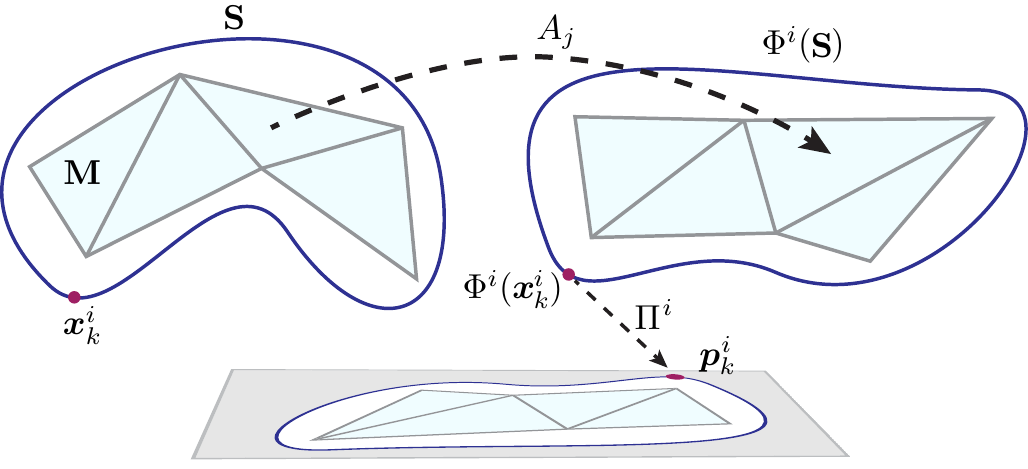}}}
\caption{Illustration of the deformation model.}
\label{fig:defmodel}
\end{figure}
We denote by $A_j = A_j(\mathbf{U})$ the linear part of each affine
transformation, linearly expressed in terms of the new vertex
locations $\mathbf{U}$. Lastly, note that subject to a deformation
$\Phi=\Phi_\mathbf{U}$ the location of the landmark points can be
simply expressed as
$\bm{\hat{x}}^i_k=\Phi_\mathbf{U}(\bm{x}^i_k)=$\aj{$\mathbf{U}\bm\alpha^i_k$}. 
This relationship along with the positional constraints are depicted
in Figure \ref{fig:defmodel}.
\vspace{-0.5em}

\subsection{Landmark-guided 3D deformation}
Our goal is to deform the template $\mathbf{S}$ such
that \eqref{eqn:pos_const_1} is satisfied without introducing local
distortions to its shape. A popular approach to prevent distortion is minimizing the as-rigid-as-possible (ARAP) functional \cite{Alexa,ARAP}:
\begin{equation}
\label{eqn:ARAP}
f_{\mathtt{ARAP}}(\mathbf{U}) = \sum_{j=1}^m ||A_j -
  R_j||^2_F|t_j|,
\end{equation}
where $R_j \in SO(3)$ is the closest rotation to $A_j$ and $|t_j|$ is
the normalized volume of the $j$'th tet. Intuitively, ARAP tries to
keep the local transformations applied to each tet of the mesh as
similar as possible to a rigid transformation. \new{Note that while
  the ARAP functional is non-convex, it is convex-quadratic for fixed rotations $R_j$.} % We take a block-coordinate descent approach where we solve for $\{R_j\}$ fixing $\{A_j\}$ and solve for $\{A_j\}$ fixing $\{R_j\}$.
% The choice of $R_j$ is discussed in section \ref{sec:realizing} .

The ARAP functional minimizes the $\ell_2$-norm of a ``non-rigidity''
measure, which strives to evenly distribute local deviations from rigid
transformation. As such, it fails to faithfully represent
articulation and local deformations. Moreover, it is not straightforward to adapt
this functional alone to benefit from having many annotated image
exemplars. \aj{In this work, we also use the ARAP functional, but allow non-uniform distribution of distortion by assigning local stiffness as described in the next section.}

%-------------------------------------------------------------------------
\section{Learning stiffness for articulation and deformation}
\label{sec:mystuff}
Natural objects do not usually deform in a uniform manner; some parts
such as joints deform a lot more while parts such as the limbs and
skull stay rigid. \aj{In order to model such deformation and articulation, we
introduce the notion of local stiffness, which specifies how
much distortion is allowed at each tet. We learn local stiffness from
data using a sparsity promoting energy,
so large deformations are concentrated in regions that require them
across many images.}

We depart from the traditional skeleton models, which are a set of rigid sticks connected by deformable
joints \cite{SCAPE,YanPollefeys}. While skeletons excel at modeling articulation, they only possess two level of
stiffness, rigid or not. In contrast, our model can represent multiple
levels of stiffness, which is essential for representing local
deformations. Moreover, using local stiffness does not require prior
knowledge, such as the number of sticks and joints.  In this section we discuss how we simultaneously deform the
template $\mathbf{S}$ to match each of the images $I_1,\ldots,I_N$
while learning the stiffness. 

\subsection{Modeling local stiffness}
Denote by $\mathbf{U}^i$ the deformation mapping $\mathbf{S}$ to the $i$'th image $I^i$, and by $\{A^i_j\}$ the linear transformations associated with its tets. Inspired by \cite{Yaron,ContSingVal}, we control deformations by explicitly imposing constraints on their linear parts. 

First we require that each $A^i_j$ satisfies
\begin{equation}
\label{eqn:const_det}
\det(A^i_j) \geq 0,
\end{equation}
which entails that the mapping is locally injective and orientation preserving; in particular, tets may not flip. Second, we bound the \emph{local isometric distortion} with the constraint
\begin{equation}
\label{eqn:const_bi}
\max\left\{\|A^i_j\|_2,\|{A^i_j}^{-1}\|_2\right\}\leq1+\eps+{s}_j
\end{equation}
where $\|\cdot\|_2$ is the operator (spectral) norm. \new{The small
  constant $\eps\geq0$ is common for all tets and governs the degree of global non-rigidity. $s_j \geq 0$ is the local \emph{stiffness} for the $j$'th tet controlling how much this particular tet may deform.} Note that $\eps$ and ${s}_j$ are not image specific (i.e. they are
independent of $i$) and encode the class-prior of how an object class
can deform and articulate. 

Intuitively, $\|A^i_j\|_2$ and $\|{A^i_j}^{-1}\|_2$ quantify the
largest change of Euclidean length induced by applying
$A^i_j$ to any vector. Therefore, Equation \eqref{eqn:const_bi} bounds local
length changes by a factor of $1+\eps+{s}_j$. If, for example,
$\eps= {s}_j=0$ then $A^i_j$ must be a rotation; looser bounds
allow ``less locally isometric''  deformations. In practice, $\eps$ is
set to a small value and is fixed throughout the experiments.

\subsection{Optimizing articulation and deformation}
Subject to these constraints, we propose minimizing an energy comprising three terms:
\begin{equation}
\label{eqn:func_sum}
f = f_{\mathtt{DEFORM}} + \lambda f_{\mathtt{POS}}  + \eta f_{\mathtt{STIFFNESS}}.
\end{equation}
$f_{\mathtt{DEFORM}}$ is defined via the ARAP deformation energy \eqref{eqn:ARAP} as
\begin{equation}
\label{eqn:func_deform}
f_{\mathtt{DEFORM}} = \frac{1}{N}\sum_{i=1}^N f_{\mathtt{ARAP}}(\mathbf{U}^i).
\end{equation}
$f_{\mathtt{POS}}$ is defined by 
\begin{equation}
\label{eqn:func_pos}
f_{\mathtt{POS}} = \frac{1}{N}\sum_{i=1}^N \sum_k
\bigg\|\begin{bmatrix}\bm{p}^i_k\\1\end{bmatrix}
-
\Pi^i\begin{bmatrix}\mathbf{U^i}\bm\alpha^i_k\\1\end{bmatrix}
\bigg\|_2^2,
\end{equation}
which accounts for the user prescribed correspondences and the camera parameters, aiming to satisfy \eqref{eqn:pos_const_1}. Lastly, we set
\begin{equation}
\label{eqn:func_stiffness}
f_{\mathtt{STIFFNESS}} = \|\bm{s}\|_1,
\end{equation}
where $\bm{s}$ is the vector whose elements are the local stiffness bounds $\{{s}_j\}$. This L1 regularization encourages most $s_i$ to be 0, so that only those tets that must distort are allowed to do so.

$\lambda$ is a parameter that controls the trade-off between
satisfying the constraints and preserving the original shape of
$\mathbf{M}$. $\eta$ is a parameter that controls the strength of the
stiffness regularization. As $\eta$ increases, it forces most $A_j$ to stay
rigid and as $\eta$ approaches 0 the solution approaches that of the ARAP
functional and the positional constraints. See Section~\ref{sect:exp_detail} for parameter settings.

In conclusion, jointly deforming the template $\mathbf{S}$ to match
each of the images $I_1,\ldots,I_N$, while estimating the local stiffness boils down to the following optimization problem:
\begin{align}
\label{eqn:opt_nonconvex}
  \min_{\{\mathbf{U}^i\},\{\Pi^i\},\bm{s}}\quad & f_{\mathtt{DEFORM}} + \lambda f_{\mathtt{POS}}  + \eta f_{\mathtt{STIFFNESS}}\\
  \text{s.t.}\quad & A_j^i = A_j^i(\mathbf{U}^i), \quad\forall \;
  j={1,\dots,m},\; i=1,\dots,N\nonumber\\
  & \det(A^i_j) \geq 0,\nonumber\\
  & \max\left\{\|A^i_j\|_2,\|{A^i_j}^{-1}\|_2\right\}\leq1+\eps+{s}_j,\nonumber\\
  & {s}_j \geq0. \nonumber
\end{align}

Note that usually in prior work, deformations are solved independently for each
set of positional constraints, since there is nothing that ties
multiple problems together. Introducing a shared stiffness
field allows us to leverage information from multiple images and improve the quality of results for all images.

\subsection{Realizing the optimization}
\label{sec:realizing}
Optimizing \eqref{eqn:opt_nonconvex} is not straightforward, as it involves the non-convex constraint \eqref{eqn:const_bi}. We realize these constraints in a convex optimization framework based on the construction presented in \cite{ContSingVal} for optimization subject to bounds on the extremal singular values of matrices.

This \new{previous work} makes the observation that the set of matrices
whose maximal singular value, $\sigmax$, is bounded from above by some constant $\Gamma\geq0$ is
 convex and can be written as a linear matrix inequality (LMI): 
\begin{equation}
  \label{eq:above}
  \mathcal{C}^{\Gamma} = \left\{ A\in \mathbb{R}^{n\times n} :
      \begin{pmatrix}
        \Gamma I & A \\
        A^T & \Gamma I
      \end{pmatrix} \succeq 0 \right\}.
\end{equation} 
It is further shown that for any rotation matrix $R\in SO(n)$, the set
\begin{equation}
  \label{eq:below}
R\mathcal{C}_\gam = \left\{ RA \in \mathbb{R}^{n \times n} | \frac{A +
    A^T}{2} \preceq \gam I\right\},
\end{equation}
is a maximal convex subset of the non-convex set of matrices with
non-negative determinant whose minimal singular value, $\sigmin$, is
bounded from below by some constant $\gamma\ge0$. \new{This calls for
  an iterative algorithm in which $R$ is updated in each iteration so
  as to explore the entire set of matrices with bounded minimum
  singular value. 
As suggested by \cite{ContSingVal}, a natural choice for $R$ is the
closest rotation to $A$. This choice, in turn, also minimizes the ARAP
functional in Equation \eqref{eqn:ARAP} for a fixed $A$.}

In order to employ the convex optimization framework of \cite{ContSingVal}, we rewrite the constraints \eqref{eqn:const_det} and \eqref{eqn:const_bi} as
$$
1/c_j \leq \sigmin(A^i_j) \leq \sigmax(A^i_j) \leq c_j \ \ \textrm{and} \ \ \det(A^i_j) \geq 0,
$$
with $c_j = 1+\eps+{s}_j$. This follows by observing that $\|A^i_j\|_2 = \sigmax(A^i_j)$ and $\|{A^i_j}^{-1}\|_2 = 1/\sigmin(A^i_j)$. Plugging \eqref{eqn:opt_nonconvex} into the framework of \cite{ContSingVal} then yields the following optimization problem:
\begin{align}
\label{eqn:opt_convex}
  \min \quad & f_{\mathtt{DEFORM}} + \lambda f_{\mathtt{POS}}  + \eta f_{\mathtt{STIFFNESS}} \\
  \text{s.t.}\quad & A_j^i = A_j^i(\mathbf{U}^i), \quad\forall \;
  j={1,\dots,m},\; i=1,\dots,N\nonumber\\
  & A_j^i \in \mathcal{C}^{\Gamma^i_j},\nonumber\\
  & A_j^i \in R^i_j\mathcal{C}_{\gamma^i_j},\nonumber\\
    & {{s_j} \ge 0}, \nonumber \\
  & \Gamma^i_j \leq {(1+\eps+{s_j})} ,\nonumber\\
  & {\frac{1}{(1+\eps+{s_j})}} \leq \gamma^i_j , \nonumber
\end{align}
whose optimization variables are $\{\mathbf{U}^i\}$,$\{\Gamma^i_j\}$,$\{\gamma^i_j\}$ and $\bm{s}$.

Lastly, we note that the last constraint of \eqref{eqn:opt_convex} is convex; in fact, following a standard derivation (e.g., see \cite{Boyd}), it can be equivalently rewritten as the convex second-order cone constraint
$$
\sqrt{4+(\gamma_j-(1+\eps+{s_j}))^2} \leq \gamma_j+(1+\eps+{s_j}).
$$
Therefore, \new{with fixed $\{R^i_j\}$ and $\{\Pi^i\}$}, Equation
\eqref{eqn:opt_convex} is a semidefinite program (SDP) and can be
readily solved using any SDP solver. \new{However, note that the entire problem is not
  convex due to the interaction between $R^i_j, \mathbf{U}^i$, and
  $\Pi^i$. Thus, we take a block-coordinate descent approach where we
  alternate between two steps: (a) update $R^i_j$ and $\Pi^i$ fixing
  $\mathbf{U}^i$, (b) update $\mathbf{U}^i$ fixing $R^i_j$ and $\Pi^i$
  via solving Equation \eqref{eqn:opt_convex}. As in \cite{ContSingVal}, we
  find that allowing the surface to deform gradually makes the
  algorithm less susceptible to local minima. To this end, we
  initialize the procedure with a large $\eta$, which controls the
  degree of non-rigidity, and slowly reduce its value as the algorithm converges. This algorithm is outlined in Algorithm \ref{alg}.}

% As suggested in \cite{ContSingVal}, we follow an alternating
% optimization scheme whereby we iterate the following steps: (a)
% optimize the SDP \eqref{eqn:opt_convex}; (b) update the rotation
% matrices $R^i_j$ to be the closest rotations to current estimates of
% $A^i_j$; and (c) update the projection matrices $\Pi^i$ with respect
% to the current estimate of  $\mathbf{U^i}$ using bundle adjustment.
\begin{algorithm}
\new{
\DontPrintSemicolon % otherwise, \For(\tcp*{}) inserts ';' after "do"..
\KwIn{Template 3D mesh $\mathbf{S}$, its auxiliary tetrahedral mesh $\mathbf{M} = (\mathbf{V}, \mathbf{T})$, and $N$ 3D-to-2D annotated images $\{I^i\}$}
\KwOut{$N$ deformed auxiliary tetrahedral meshes vertices $\{\mathbf{U}^i\}$, the projection matrices $\{\Pi^i\}$, and the stiffness model $\bm{s}$}
$\mathtt{maxIter} = 10$\string;\;
$\mathbf{\widetilde{U}^i} = \mathbf{V},\quad i = 1\dots N$\,;  \tcp*{initialize}
% \For{$\eta \leftarrow \eta_{\max}$  \KwTo  $\:\eta_{\min}$}{ 
\For(\tcp*{\vspace{-1em}warm start}){$\eta \leftarrow \eta_{\max}$ \KwTo $\:\eta_{\min}$ }{
  $\mathbf{U^i}^{(0)} = \mathbf{\widetilde{U}^i}$\string;\;
  $t = 0$\string; \;
  \Repeat{ convergence \normalfont{or} $t > \mathtt{maxIter}$} {
    % Compute ${\Pi^i}^{(t)}$ by solving Equation \eqref{eqn:pos_const_1} via direct linear transform + bundle adjustment \cite{HZ}\;
    Compute ${\Pi^i}^{(t)}$ by solving Equation \eqref{eqn:pos_const_1} with ${\mathbf{U}^i}^{(t)}$\string; \;
    Compute the polar decompositions ${A_j^i}^{(t)} = {R_j^i}^{(t)}{S_j^i}^{(t)}$\string; \;
    Update $\{{\mathbf{U}^i}^{(t+1)}\}, \bm{s}^{(t+1)}$ by solving Equation \eqref{eqn:opt_convex} with ${\Pi^i}^{(t)}$ and ${R_j^i}^{(t)}$\string; \;
    $t = t + 1$\string; \;
  }
  $\mathbf{\widetilde{U}^i} = \mathbf{U^i}^{(t)}$\string; \;
}
\KwRet{$\{{\mathbf{U}^i}^{(t)}\}, \{{\Pi^i}^{(t)}\}, \bm{s}^{(t)}$}
}
\caption{\new{Jointly solving for the deformations and the stiffness \label{alg}}}
\end{algorithm}
\vspace{-0.8em}

\section{Experimental Detail}
\label{sect:exp_detail}
We use our approach as described to modify a template 3D mesh
according to the user-clicked object pose in 2D images. \new{We first
  compare our approach with the recent method of Cashman et
  al. \cite{Cashman}, which is the closest work to ours with publicly
  available source code \cite{dolphin}. We then present an ablation study where key
  components of our model are removed in order to evaluate their
  importance and provide  qualitative and quantitative evaluations.} % We   without any distortion
% bounds (i.e. removing \eqref{eqn:const_bi}) and with constant distortion
% bounds (i.e. fixing $s_j$ to a constant) to compare with the
% complete system with learned stiffness. 

We experiment with two object categories, cats and horses. We
collected 10 cat and 11 horse images \aj{in a wide variety of poses} from the
Internet. Both of the template 3D meshes were obtained from the Non-rigid World dataset
\cite{Bronstein}. \aj{These templates consist of $\sim$3000 vertices
and $\sim$6000 faces, which are simplified and converted into tetrahedral meshes of 510, 590 vertices and 1500,
1700 tets for the cat and the horse respectively via a tet generation
software \cite{tetgen}. We manually simplify the mesh in MeshLab until
there are around 300 vertices. We found automatic simplification
methods over-simplify skinny regions and fine details, leading to a poor volumetric
tet-representation.}  The cat template and its auxiliary tetrahedral mesh are shown in
Figure \ref{fig:surftet}. The template mesh used for horses can be
seen in Figure \ref{fig:seg}.
For all experiments we set $\eps=0.01$, and $\lam = 10$. \new{In order
  to allow gradually increasing levels of deformation, we use $\eta_{\max} = 0.5$ and $\eta_{\min} = 0.05$ with 10 log-steps in between for all experiments}. % This warm start also improves the camera projection parameters as it avoids getting stuck in
% a local minimum early on.
The values for $\eta$ and $\lam$ were set by hand, but deciding on the values did not require much tuning.

In each iteration, the camera parameters are computed using the 2D-to-3D
correspondences. We initialize the parameters using the direct linear
transform algorithm and refine it with the sparse bundle adjustment package
\cite{HZ,SBA}. In order to obtain annotations, we set up a simple
system where the user can click on 2D images and click on the corresponding 3D points in the
template mesh.  Our system does not require the same vertices to be
annotated in every image. The average number of points annotated
for each image for both cats and horses was \aj{29 points}. 

\section{Results}
\noindent\textbf{Comparison with \cite{Cashman}}
Cashman et al. employ a low resolution control mesh on the order of
less than 100 vertices which is then interpolated with Loop subdivision. In order to apply their method to ours, we simplified our template mesh with quadratic
decimation until we reach around ~150 vertices while retaining the key features of the template mesh as
much as possible (shown in inset). 
\setlength{\columnsep}{2pt}
\begin{wrapfigure}[4]{o}{0.35\columnwidth}
% \hfill
\vspace{-1.2em}
\begin{center}
\includegraphics[width=.35\columnwidth]{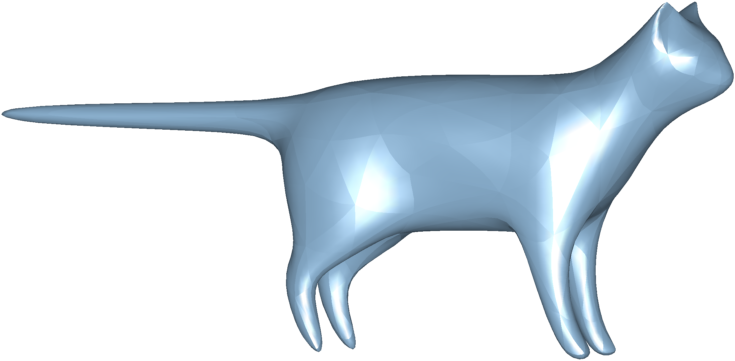}
\end{center}
\label{fig:dolphintemplate}
\end{wrapfigure}
Since their method relies on silhouettes, we provide hand-segmented
silhouettes to their algorithm along with the user-clicked points. We transferred the user-clicks from the
full mesh to the simplified mesh by finding the closest 3D
vertex in the simplified mesh for each labeled vertex in the full
resolution mesh. We did not include points that did not have a close enough 3D vertex due to
simplification. On average 24 points were labeled for their
experiment and we use their default parameters.

\begin{figure}[h]
\begin{center}
  \includegraphics[width=\linewidth]{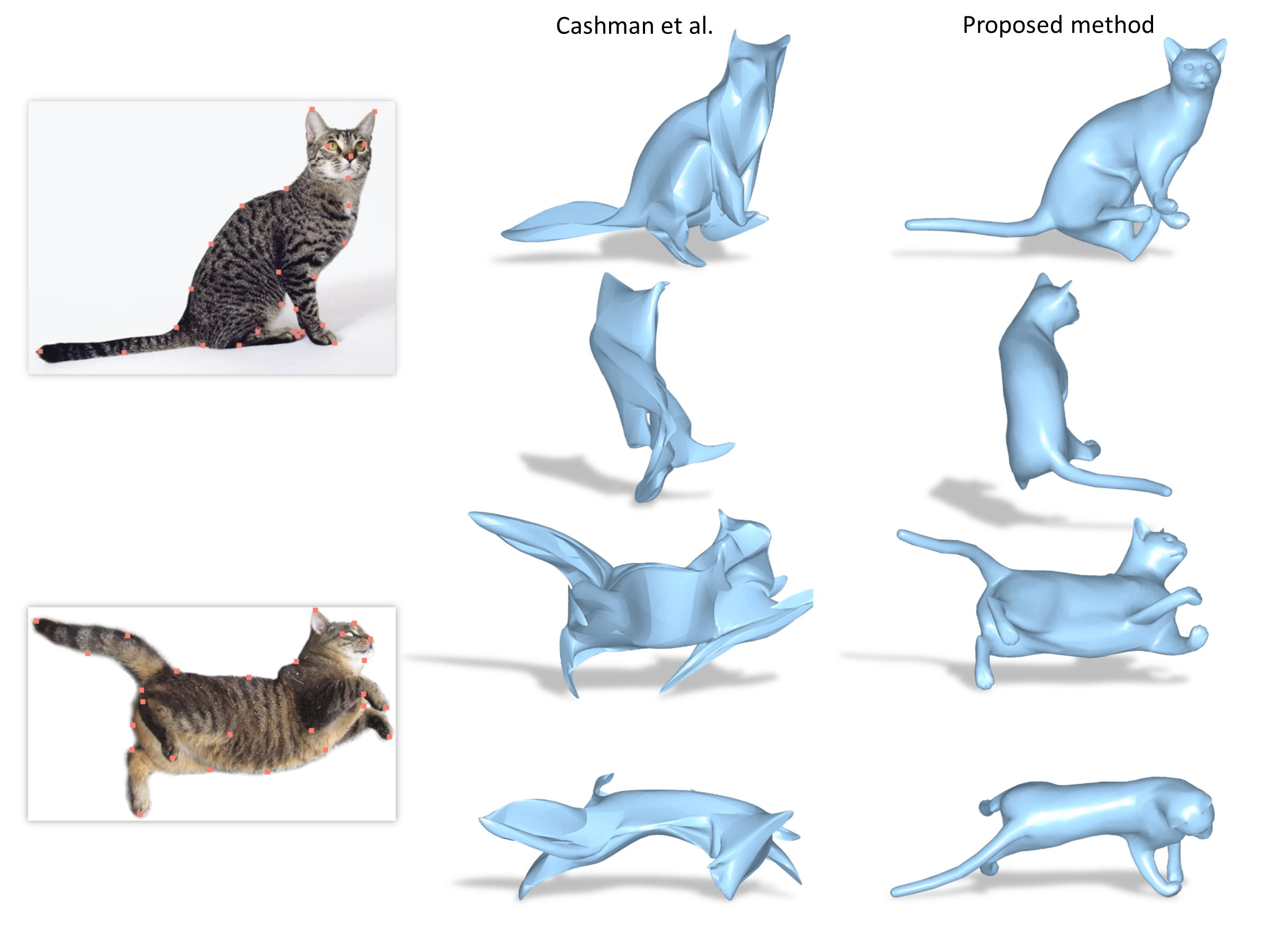}
  \caption{Comparison with \cite{Cashman}: the first column shows the
    user-clicked input images, the second column shows the result of
    \cite{Cashman} and the third column shows the result of
    our proposed method. Two views are shown for each image, one from the
  final estimated camera and another from an orthogonal view point. Our
  method is more robust to large deformations and retains the volume of the model. Note that silhouettes, along with the user-clicked points, are used to obtain the results for \cite{Cashman}.}
\label{fig:dolphin}
\end{center}
\end{figure}
\begin{figure*}[ht]
\centerline{{\includegraphics[width=\linewidth,height=160mm]{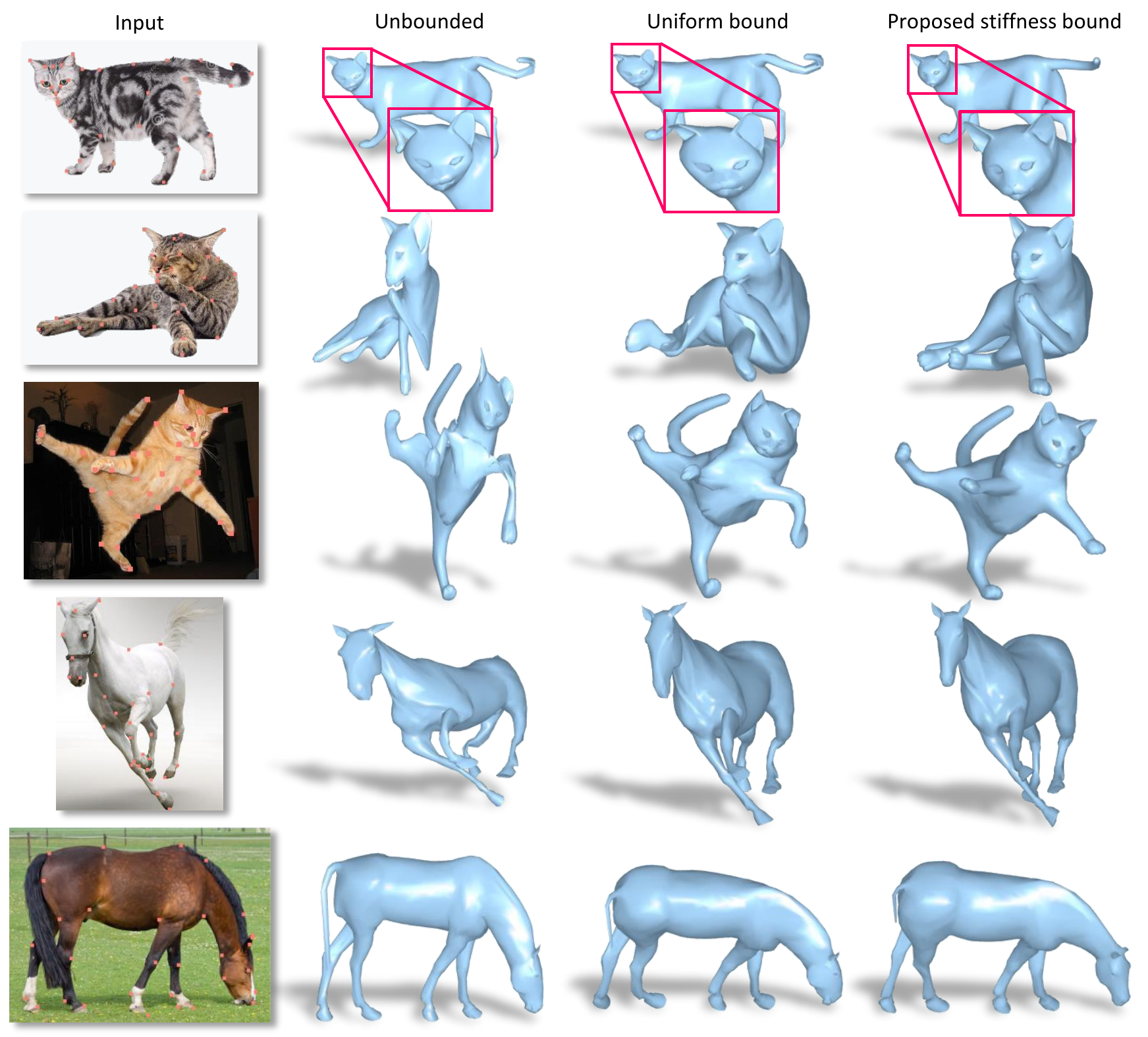}}}
   \caption{Comparison of proposed approach vs results with its key
     components removed. User-clicked input images (first column). Unbounded (second column) is the model without any bounds
     on the distortion leaving just the volumetric ARAP
     energy. Uniform (third column) is when the stiffness bounds ($s_j$ in Equation \eqref{eqn:const_bi}) are replaced by a single constant, which is the approach of \cite{ContSingVal} applied to 2D
     positional constraints. The last column shows the results with our complete framework where the stiffness bounds and the deformations are jointly optimized. Without the animal-specific stiffness, distortions either spread out too much or concentrate around the positional constraints.}
\label{fig:big}
\end{figure*}

Figure \ref{fig:dolphin} compares the results obtained with the method of
\cite{Cashman} and our model. Two views are shown for each result, one
from the estimated camera pose and another from an orthogonal viewpoint. As the authors
in \cite{Cashman} point out, their method focuses on modeling shape and is not designed for highly articulated
objects such as cats. Consequently, we can see that it
has difficulties dealing with the wide range of poses present in our
cat dataset. Regions such as limbs and tails especially lose
their original shape. Their method is based on surface deformation, which does not have a notion
of volume. This causes flattening of the 3D models as can been seen in the
orthogonal views. Since we guarantee worst-case distortion and
orientation preserving deformation of the auxiliary mesh, our surface
reconstructions are well behaved compared to \cite{Cashman}. Recall that silhouettes, along with the user-clicked points, are used to obtain the results for \cite{Cashman}.

% \subsection{Ablation Study}
% \label{sec:ablation-study}
\noindent \textbf{Qualitative evaluation} The 3D models in Figure
\ref{fig:pipeline} were obtained using our proposed framework.
We now evaluate the effectiveness of the proposed framework by comparing the results without any distortion bounds (i.e. removing Equation
\eqref{eqn:const_bi}) and with constant distortion bounds (i.e. fixing
$s_j$ to a constant). Qualitative results of this ablation study are shown in Figure \ref{fig:big}. The first column shows input images along with
their user-clicked points. The second column shows results obtained with no bounds, leaving just the ARAP energy, which we refer to as 
\texttt{Unbounded}. This is similar to the approach used in \cite{Kholgade}, but with volumetric instead of surface deformation. The third column, \texttt{Uniform}, shows results obtained with
a uniform bound, where the stiffness $1 + \eps + s_j$ is replaced with
a single constant $c_j = 2$ for all faces. This is the deformation
energy used in \cite{ContSingVal} applied to 2D positional constraints. The constant was slowly increased from 1 to 2 in a manner
similar to $\eta$ in order to allow for increasing levels of deformation. Finally, in the last column we show results obtained with the proposed framework
where the distortions are bounded with local stiffness. 

% We evaluate the effectiveness of the proposed framework by comparing
% the results without any distortion bounds (i.e. removing Equation
% \eqref{eqn:const_bi}) and with a constant distortion bound (i.e. fixing
% $s_j$ to a constant). \new{When no bounds are used, we are left with
%   just the ARAP energy and we refer to this method as \texttt{Unbounded}. This is similar to the approach of
%   \cite{Kholgade}, but with volumetric instead of surface
%   deformation}. \new{For a constant distortion bound, we replace the stiffness $1
%   + \eps + s_j$ with a single constant $c_j = 2$ for all faces and
%   refer to this method as \texttt{Uniform}}. This is the deformation
% energy used in \cite{ContSingVal} applied to 2D positional
% constraints. The constant was slowly increased from 1 to 2 in a manner
% similar to $\eta$ in order to allow for increasing levels of deformation.

\aj{First, notice the wide range of poses present in the images used; some
are particularly challenging requiring large deformation from the template 3D
mesh. In general, \texttt{Unbounded} concentrates high
distortions near positional constraints causing unnatural
stretching and deformation around limbs and faces. This is
evident with horse legs in row 4 as \texttt{Unbounded}
deforms them in an elastic manner. \texttt{Uniform} distributes the distortions, however, when the
pose change from the template is significant, distortions spread out
too much causing unrealistic results as seen in rows 2 and 3. The unnatural distortion of the faces is
still a problem with \texttt{Uniform}. The proposed framework alleviates
problems around the face and the horse limbs as it learns that those
regions are more rigid.%  The local deformation around the neck of the grazing horse in row 5 is well captured by
% our stiffness model; this may be challenging for skeletal models unless many
% sticks and joints are used.
} Please refer to the supplementary materials for comprehensive results of all cat and horse experiments.

\begin{figure}[]
\centering
  \includegraphics[width=\linewidth,height=40mm]{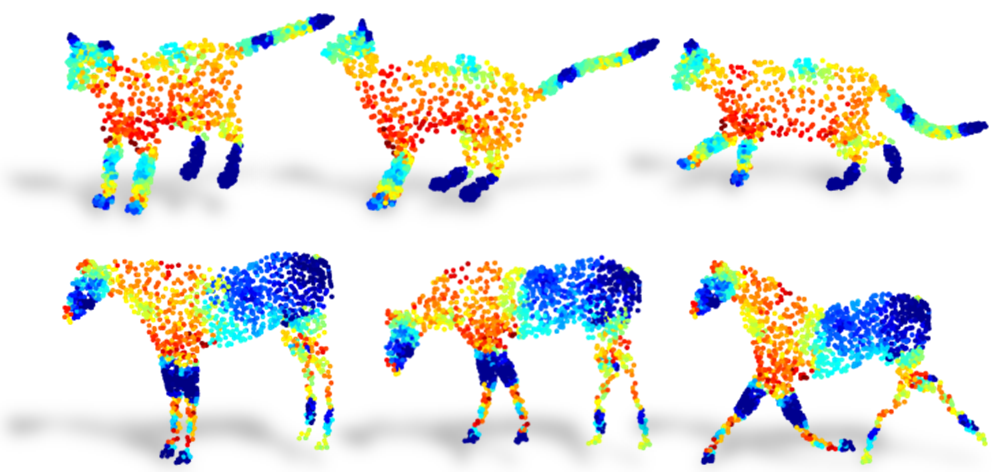}
  \caption{Visualization of the learned stiffness values. Blue indicates rigid regions while red indicates highly deformable
    regions.}
\label{fig:stiffness}  
\end{figure}

\begin{figure}[]
\centering
  \includegraphics[width=\linewidth,height=40mm]{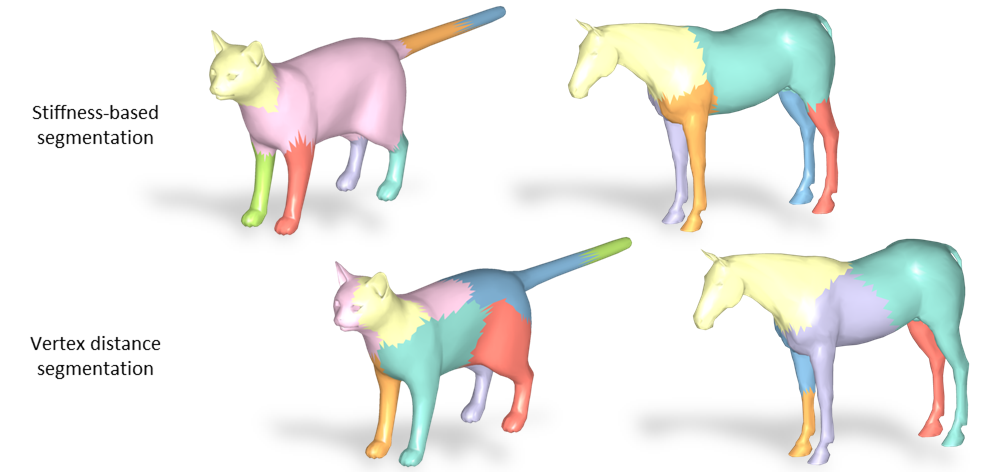}
  \caption{Another visualization of the learned stiffness by means of segmentation. Segmenting the template mesh using stiffness illustrates regions that deform together as learned by our framework. We see that they correspond to semantically reasonable segmentations. We show segmentation results based on vertex distance alone as a comparison. 
}
\label{fig:seg}  
\end{figure}

We provide visualizations of the learned stiffness model in Figure \ref{fig:stiffness} and \ref{fig:seg}. Figure \ref{fig:stiffness} visualizes the learned stiffness values for
cats and horses in various poses. We show the centroid of tetrahedra faces colored by their stiffness values in log scale. Blue indicates rigid
regions while red indicates highly deformable regions. Recall that there is one stiffness model for each animal category. The level of
deformations present in the input images are well reflected in the
learned stiffness model. For cats, the torso is learned to be highly
deformable allowing the animal to twist and curl, while keeping the
skull and limbs more rigid. Similarly for horses, the neck, the
regions connecting the limbs as well as the joints are learned to be
deformable while keeping skull, limbs, and buttocks region
rigid. \mew{The fact that the buttocks is considered rigid is
anatomically consistent, since horses have a rigid spine making them suitable for riding \cite{jeffcott1980natural}}. % maybe counter-intuitive anatomically, however, none of the 11 horse images we used display deformations around those regions.


We also present segmentation results using the learned stiffness values as another form of visualization in Figure \ref{fig:seg}. In order to obtain the segmentations, we first transferred the stiffness values from tetrahedra faces to vertices by taking the mean stiffness of faces a vertex is connected to. Then, we constructed a weighted graph on the vertices based on their connectivity, where the weights are set to be the sum of the Euclidean proximity and the similarity of the stiffness values. % The weight of the edge connecting vertex $i$ and $j$ is set to be $w(i,j) = exp^{-d(i,j)/\sig_x}+exp^{-(||s_i - s_j||/\sig_s)}$. 
We apply normalized cuts to partition this graph and interpolate the result to the surface mesh vertices using the parameterization described in Section \ref{sec:param}. We also show the segmentation results using just the Euclidean proximity as a comparison. Stiffness-based segmentation illustrates that regions which deform together as learned by our framework correspond to semantically reasonable parts. 

\begin{figure}[t]
\begin{center}
  \includegraphics[width=\linewidth,height=30mm]{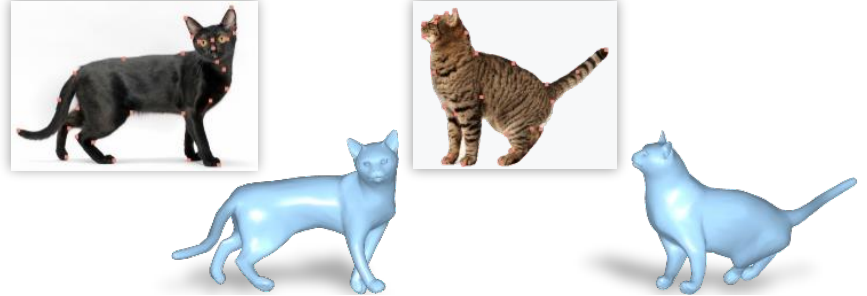}
  \caption{Deformation results using the learned stiffness from 10 cats as a fixed prior for new images.}
\label{fig:reuse}
\end{center}
\vspace{-1em}
\end{figure}

The learned stiffness model can be used as a prior to solve for
stiffness-aware deformations of new annotated images. Figure
\ref{fig:reuse} shows the results of deforming the template to new
input images via using the stiffness values learned from the previous
experiment, i.e. the new images were not used to learn the stiffness. Similar to other experiments, we do warm start where the stiffness bounds are
linearly increased from 1.01 to their actual value in 10 steps. The
results are visually very similar to the results obtained when the
stiffness was learned with those images along with the other 10 cat
images. From this perspective, the joint optimization for the stiffness and the
  deformations using multiple images is the ``training'' (Figure \ref{fig:big}), while the single-image optimization with a
  fixed stiffness prior is the ``testing'' (Figure \ref{fig:reuse}).  

\noindent \textbf{Quantitative evaluation} \new{Lastly, we conduct an evaluation against
 the ground truth by using pictures of a rendered 3D model as the input to our
  framework. Specifically, we use the TOSCA dataset
  \cite{bronstein2008numerical}, which has 11 vertex-aligned
  models of cats in various poses. We take the neutral pose (\texttt{cat0}) as the
  template and randomly project the other 10 models to produce images
  where the ground truth shape is known. We randomly sample 35 points
  and use them as the 3D-to-2D correspondences. In order to guarantee
  that these points are well distributed, we segment the model into
  15 components and make sure that there is at least one point from
  each component. These components correspond to key parts such as the
  paws, limbs, left and right ears, tail base and tip, etc. We compare
  the results of the \texttt{No Bound}, \texttt{Uniform}, and the
  proposed approach. Using this method, we produce two images from
  each ground truth model and conduct the experiment with 20 images.}
\begin{table}
\centering
\caption{Quantitative evaluation against ground-truth. The Lower the
  better for all metrics.  \label{tab:gt}}
\begin{tabular}{llllll}
% \begin{tabularx}{\linewidth}{XXXXXX}
\hline
 &  & \multicolumn{4}{c}{\uline{Distortion error metric}\cite{yoshizawa}}\\
Methods & Mean dist & Stretch & Edge & Area & Angle \\ \hline
\texttt{Unbounded} & 0.291                   & 1.01              & 0.156                  & 0.216           & 0.159          \\
\texttt{Uniform}   & \textbf{0.281}          & 1.01              & 0.13                  & 0.198           & 0.13            \\
\texttt{Proposed}  & 0.287                    & \textbf{0.996}    &
\textbf{0.105}         & \textbf{0.181}  & \textbf{0.085}  \\
% Methods & Mean dist & Stretch & Edge & Area & Angle \\ \hline
% \texttt{Unbounded} & 0.324                   & 1.018              & 0.155                  & 0.217           & 0.161            \\
% \texttt{Uniform}   & \textbf{0.298}          & 1.015              & 0.131                  & 0.202           & 0.131            \\
% \texttt{Proposed}  & 0.31                    & \textbf{0.997}    & \textbf{0.108}         & \textbf{0.182}  & \textbf{0.088}  \\
\hline
% \end{tabularx}
\end{tabular}
\end{table}

% \begin{figure*}[t]
% \begin{center}
%   \includegraphics[]{figures//gt_exp_ppt.pdf}
%   \caption{\new{Illustration of how mean L2 distance does not reflect
%     the visual quality of the deformed models: First
%     column is the ground truth model along with the point
%     correspondences used. The first row is the input image and the
%     second row shows an alternative view from below. Results of
%     \texttt{Unbounded}, \texttt{Uniform}, and proposed method are shown in columns 2, 3 and 4 respectively. Despite the unnatural deformations,
%     \texttt{Unbounded} obtains the lowest mean L2 distance even when the
%     tail is not accounded for. }}
% \label{fig:gt}
% \end{center}
% \end{figure*}

\new{We evaluate our method using several error metrics. First, we
  measure the distortions between the ground truth and the deformed
  models, which capture how natural the deformations are. We argue
  this is the most important measure since obtaining plausible
  deformations is the main goal of our algorithm. For this we use
  the stretch, edge-length, area, and angle distortion errors as
  defined in \cite{yoshizawa} by comparing the corresponding
  triangles. Additionally, we report the mean Euclidean distance between the 3D
  vertices, which measures how close the surface of the
  deformed models are to the ground truth. While a low Euclidean distance is desirable for 3D reconstruction, we do not expect a close match everywhere due to ambiguities arising
  from a single view and sparse point constraints. In particular,
  Euclidean distance is not necessarily indicative of visual
  quality. We report the average error over all 20 input
  images. Before computing the error metrics, the deformed and ground
  truth models are aligned by a similarity transform. The results are
  shown in Table \ref{tab:gt}.} \mew{As expected, all methods attain
  comparable mean Euclidean distance to ground truth, while our
  approach obtains substantially lower errors in distortion
  metrics. This demonstrates the advantage of learning stiffness from
  multiple images, yielding a more plausible deformation model. % Note that Euclidean distance does not reflect the visual quality of the deformed models as illustrated in Figure \ref{fig:gt}. Even though the baseline approaches have unnatural deformations like flattening of the thighs and bending of
 %  the hind legs, they still have lower error in terms of Euclidean
 %  distance. Even when the tail vertices are not considered, the most
 %  visually unnatural result of \texttt{Unbounded} still obtains the
 % lowest Euclidean distance to ground truth.
}

\noindent \textbf{Implementation details} With an unoptimized MATLAB
  implementation, training with 10 images took 4 hours and testing a single
  image with a learned stiffness prior took $\sim$30 minutes. We use YALMIP \cite{yalmip} for the SDP modeling and MOSEK as the
solver \cite{mosek}. Our biggest bottleneck is the SDP optimization due to many LMI
constraints. Reducing the number of tets can significantly reduce the run-time.

\section{Limitations and Future Work}
\label{sec:discussion}
\mew{Limitations of our current approach suggest directions for future
  work. One failure mode is due to a large pose difference between the
  template and the target object, which may lead to an erroneous
  camera parameters estimate (e.g., local minima), as seen in row 5 of Figure
  \ref{fig:big}. Here, the head of the horse in the image is lowered for grazing while
  the head of the horse template is upright causing a poor initialization of the camera estimate.} \new{Using a user-supplied estimate of the viewpoint or automatic viewpoint estimation methods like \cite{Tulsiani} are possible solutions.}

\mew{Another pitfall is that some parts may be bent in
  an unnatural direction as seen around the left ankle of the horse in
  row 4 of Figure \ref{fig:big}.} \new{An interesting future direction is to
  make the distortion bounds dependent on the orientation of the
  transformation. This would allow the framework to learn that certain
  parts only deform in certain directions. }

\new{Since we only have a single view for each target object, there is
  an inherent depth ambiguity, where many 3D shapes project to the
  same 2D observations. As such, some of our deformed models do not
  have the ``right'' 3D pose when seen from a different view. } \mew{This is
  illustrated in our supplementary video that shows 360 degree views of the final models. % We like to note that baseline methods also suffer from this issue and our results is more natural even when it's not the correct 3D shape because of the learned stiffness.
One possibility is to combine our method with recent single image reconstruction approaches like \cite{Joao,VVN} that use a large image collection of
the same object class to resolve the depth ambiguity.}

\new{Our method could also be enhanced to prevent surface
  intersections or reason about occlusion (e.g. if the point is
  labeled, it should be visible from the camera). Run-time is also an
  issue for adapting the stiffness model into a real-time posing
  application.} \mew{This may be addressed by recent advancements in
  efficiently computing mappings with geometric bounds \cite{Kovalsky15}.}

\section{Conclusion}
Modifying 3D meshes to fit the pose and shape of objects in images
is an effective way to produce 3D reconstruction of Internet
images. In order to fit object pose naturally, it is essential to
understand how an object can articulate and deform, especially for
highly deformable and articulated objects like cats. In this paper we
propose a method that can learn how an object class can deform and articulate from a set of user-clicked 2D images and a 3D template
mesh. We do so by introducing a notion of local stiffness that
controls how much each face of the mesh may distort. We jointly
optimize for the deformed meshes and the stiffness values in an iterative
algorithm where a convex optimization problem is solved in each
iteration. \mew{Our experiments show that learning stiffness from
  multiple images produces more plausible 3D deformations. We hope
  this motivates further developments in the area of automatic point
  correspondence for non-rigid objects, enabling fully-automated 3D
  modeling of animals from 2D images in the near future.}

\section*{Acknowledgements}
We would like to thank Austin Myers for helpful comments and suggestions. This material is based upon work supported by the National Science
Foundation under grant no. IIS-1526234, the Israel Binational Science
Foundation, Grant No. 2010331 and the Israel Science Foundation Grants No. 1265/14.

\bibliographystyle{eg-alpha-doi}
\bibliography{cat_paper}

\end{document}